\documentclass[conference]{IEEEtran}
\usepackage{cite}
\usepackage{amsmath, amssymb, amsfonts, amsthm}
\usepackage{algorithmic}
\usepackage{graphicx}
\usepackage{textcomp}
\usepackage{xcolor}
\usepackage{algorithm}
\usepackage{algorithmic}
\usepackage[labelfont=small, textfont=small]{caption}

\graphicspath{ {./} }

\def\BibTeX{{\rm B\kern-.05em{\sc i\kern-.025em b}\kern-.08em
    T\kern-.1667em\lower.7ex\hbox{E}\kern-.125emX}}

\newtheorem{theorem}{Theorem}
\newtheorem{problem}{Problem}
\newtheorem{proposition}{Proposition}
\newtheorem*{definition}{Definition}

\DeclareMathOperator*{\argmax}{arg\,max}

\title{
    Robust Multi-Agent Task Assignment in \\
    Failure-Prone and Adversarial Environments
}

\author{
    \IEEEauthorblockN{Russell Schwartz \quad\quad Pratap Tokekar}
    \IEEEauthorblockA{
        \textit{Robotics Algorithms \& Autonomous Systems Lab} \\
        \textit{University of Maryland, College Park}
    }
}

\begin{document}
    \maketitle

    \begin{abstract}
		The problem of assigning agents to tasks is a central computational challenge in many multi-agent autonomous systems. However, in the real world, agents are not always perfect and may fail due to a number of reasons. A motivating application is where the agents are robots that operate in the physical world and are susceptible to failures. This paper studies the problem of Robust Multi-Agent Task Assignment, which seeks to find an assignment that maximizes overall system performance while accounting for potential failures of the agents. We investigate both, stochastic and adversarial failures under this framework. For both cases, we present efficient algorithms that yield optimal or near-optimal results.
    \end{abstract}
    
    \section{Introduction}
	Task assignment is a central issue in most multi-agent systems, as it determines which agents should be assigned to which tasks in order to optimize the system performance. This topic has received much attention as the interest in multi-agent systems has steadily increased. Numerous formulations of the problem have been discussed with varying constraints and objectives \cite{lee,zhang,biswas}. In the simplest form, the problem of assigning agents to tasks can be formulated as that of one-to-one bipartite graph matching which can be solved efficiently in polynomial time. When additional constraints are introduced, finding the optimal assignment often proves to be computationally intractable due to the combinatorial nature of the problem. Nonetheless, algorithms for task assignment have been designed which ensure near-optimal performance with suitable runtimes. Such algorithms have found a variety of applications including vehicle-target assignment \cite{arslan}, sensor coverage problems \cite{marden}, and discrete resource allocation \cite{yamaguchi}. Many of these approaches are predicated on the assumption that agents will always complete their assigned tasks, which limits their applicability to settings where agent failures are rare or impossible. To account for uncertainty in the agents' performance, more complex models are required such as the cost-based system presented in \cite{timotheou}.

    In this paper, we focus on scenarios where the agents operate in failure-prone or adversarial environments. In such environments, an agent is susceptible to internal or external factors which can impede its performance. In particular, we consider failures and attacks which lead to complete malfunction (i.e., total failure in completing the assigned task) of individual agents. Our goal is to provide planning and coordination algorithms that are robust to such failures.
	
    We introduce the problem of robust multi-agent task assignment in both stochastic and adversarial environments. By stochastic, we refer to scenarios in which agents fail independently with probability $p$ (where $p$ is known) as the result of some random process. By adversarial, we refer to scenarios in which at most $\alpha$ agents fail (where $\alpha$ is known) as the result of an attack by an intelligent adversary. The latter is, in essence, a Stackelberg game in which the assignment represents the leader's strategy, and the attack represents the follower's strategy \cite{myerson}. For both environments, we provide exact and approximate solutions, with a particular effort placed on achieving runtimes that are independent of the number of agents. This makes such algorithms suitable for applications where the number of agents is large. We conclude with empirical results of our algorithms' performance and a brief discussion of the possible applications of these concepts both within the field of robotics and elsewhere.
    
    \begin{figure}[t]
		\centering
		\includegraphics[width=0.8\linewidth]{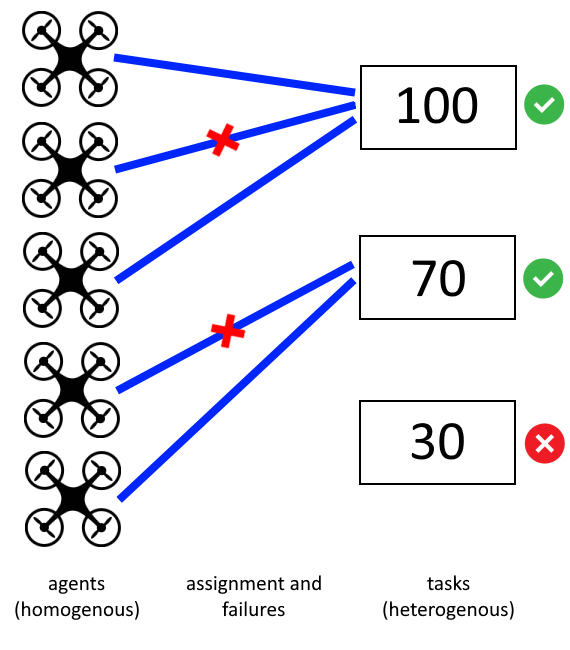}
		\caption{An example of our agent/task assignment framework. Five agents are assigned to three tasks. Two of the agents fail and one of the tasks is not completed, resulting in a total profit of 170.}
	\end{figure}
	
	\subsection{Related Work}
	There exists a large body of work on task assignment and related control problems, with emphasis on accounting for various spatial and temporal constraints. Despite the attention the subject has received, there is relatively little literature on failure-prone systems. This is in part due to the added complexity that comes with modeling agent failures. 
	
	Stochastic environments are most commonly addressed, since they appear often in the real world. Heuristics have been developed for various applications that help to account for random failures, but do not necessarily achieve optimality \cite{timotheou,alami}. Our work is concerned with a particular framework in which optimality (in terms of expected value) can be achieved. 
	
	Assignment in adversarial environments, on the other hand, have not yet been as well explored. Aside from the obvious military applications, adversarial environments effectively models a variety of scenarios in which worst-case behavior is of key concern \cite{patterson,mohan}. Most closely related to our work, the notion of adversarial failure has been successfully used to model agent failures in the field of active target tracking \cite{zhou}. We investigate the same model used in that paper but in the context of task assignment. 
	
    Similar problems have been addressed, specifically, the min-max knapsack, in which the values of the items in the knapsack are uncertain \cite{yu}. Our model corresponds to knapsack optimization where the \emph{weights} are uncertain. As far as we are aware, the particular combinatorial optimization problems involved in our models are unique to this work.

	\subsection{Framework and Notation}
	For the purpose of this paper, the Robust Multi-Agent Assignment Problem (RMAAP) is concerned with the following natural framework.

	A set of tasks is given, each of which has an associated profit. A number of agents is also given, each of which can be assigned to at most one task. A task is completed successfully if it has at least one successful agent assigned to it. The goal is to determine the assignment of agents to tasks that maximizes the sum of the profits of the completed tasks. 

	\begin{itemize}
		\item Let $k$ denote the number of tasks and $N$ the number of agents.
		\item Let $T = [t_1, t_2, \cdots, t_k] \in \mathbb{R}^k$ denote the profits of the tasks. For convenience, we assume that $T$ is given in \emph{decreasing} order ($t_i \ge t_j$ for all $1 \le i < j \le k$).
		\item Let $x = [x_1, x_2, \cdots, x_k] \in \mathbb{Z}_{\ge 0}^k$ denote the assignment that assigns $x_i$ agents to task $i$. An assignment $x$ is valid iff $\sum_{i=1}^{k}x_i \le N$. 
		\item Let $H(n) = \min(1, n)$ denote the shifted heaviside function. This represents the completion indicator variable for a task with $n$ successful agents assigned to it.
	\end{itemize}

	When no failures may occur, the total profit under assignment $x$ is given by $\sum_{i = 1}^k t_i \cdot H(x_i)$. In this case, the task assignment problem is trivial: total profit is clearly maximized by assigning one agent to each of the $\min(k, N)$ most valuable tasks. The problem becomes much more complex when failures are possible.
	
	\subsection{Organization and Contributions}
	The organization of the remainder of the paper is as follows. In Section 2, we formulate RMAAP for stochastic environments and then provide two different exact solutions (one pseudo-polynomial that runs in $O(k + N \log k)$ time and one strongly polynomial that runs in $O(k^2)$ time). In Section 3, we re-formulate RMAAP for adversarial environments and provide a justification for the attack model. We then give an exponential-time exact solution that runs in $O\Big(\frac{N^k}{(k-1)!^2}\Big)$ time along with a proof of NP-Hardness. This is followed by a polynomial-time approximate solution that runs in $O(k^2)$ time. We conclude with empirical results for the optimality of this solution and compare it to various baselines, followed by a discussion of future work.

	\section{Stochastic Resilience}
	\subsection{Problem Formulation}
	For illustration, consider a scenario in which you wish deploy a set of autonomous robots (agents) to survey a set of objects in an area (tasks), where the importance (associated profit) of each object has been determined ahead of time. You know that each robot has a chance of suffering a mechanical malfunction that will prevent it from completing its assignment. As before, each object only requires one successful robot to be accurately surveyed (i.e., there is no benefit gained from having more than one). Assuming these malfunctions occur independently with probability $p$, what assignment will maximize the \emph{expected value} of the total profit? This scenario is modeled by the following non-linear mixed-integer problem.
	
	\begin{problem}[Stochastic Failure]
		Given task values $T \in \mathbb{R}^k$, number of agents $N \in \mathbb{Z}_{\ge 0}$, and failure probability $p \in [0, 1]$, find an assignment $x \in \mathbb{Z}^k$ that
		\begin{align*}
			\text{maximizes\quad}& \sum_{i = 1}^{k} t_i(1 - p^{x_i})\\
			\text{subject to\quad} &\sum_{i = 1}^{k} x_i \le N \text{\quad and\quad} x_i \ge 0 \hspace{0.8mm}\text{ for all }\hspace{0.8mm} 1 \le i \le k
		\end{align*}
	\end{problem}

	Note that the objective function follows directly from the expected value for $H(x_i)$. The key notion introduced here is that of redundancy. It may now be advisable to assign multiple agents to the same tasks in the hopes that if one agent fails, the rest will be able to take its place. This comes at the cost of leaving the less valuable tasks unattempted. For example, given $N = 3$, $T = [70, 30, 10]$, $p = 0.3$, the optimal solution is $x^* = [2, 1, 0]$ with expected profit of 84.7.
	
	\subsection{Exact Solution}
	We begin by presenting an intuitive greedy algorithm that solves Problem 1 exactly in pseudo-polynomial time. Since its runtime is dependent on the number of agents, we then refine our algorithm by using continuous relaxation to achieve strongly-polynomial runtime.
	
	A greedy algorithm will give the optimal solution to Problem 1. Intuitively, we iterate over the agents, assigning them sequentially to the task that results in the largest marginal increase in expected value. By using a max heap to keep track of the most profitable tasks, we can implement this approach in $O(k + N \log k)$:
	
	\begin{algorithm}[h]
		\caption{Greedy solution to Problem 1}
		\begin{algorithmic}
			\STATE $x \gets [0, 0, \dots, 0]$
			\FOR{$i = 1$ to $N$}
				\STATE $z \in \argmax_{1 \le j \le k}\big(t_j \cdot p^{x_j}(1 - p)\big)$
				\STATE $x_z \gets x_z + 1$
			\ENDFOR
			\RETURN $x$
		\end{algorithmic}
	\end{algorithm}

	Since the runtime is dependent on $N$, the greedy algorithm is pseudo-polynomial. This makes it satisfactory for only sufficiently small values of $N$. For instance, if we wish to coordinate a swarm of $10^5$ robots, the runtime of Algorithm 1 could easily exceed our time limit even when there are relatively few tasks. Alternatively, the problem can be solved exactly in strongly polynomial time by first solving the continuous relaxation (i.e., allowing for fractional agents) and then converting the solution back to the discrete case. 
	
	We modify Problem 1 to allow $x_i \in \mathbb{R}$ and denote the objective function $f$. Also note that since an optimal solution will never leave agents un-assigned, we can assume that $\sum x_i = N$. Now, we use the method of Lagrange multipliers to deduce that an optimal assignment must form a solution to the following system of equations \cite{bertsekas}:

	\begin{align*}
		&\frac{\partial f}{\partial x_1} = \frac{\partial f}{\partial x_2} = \cdots = \frac{\partial f}{\partial x_k} \\
		\implies& -\ln(p)t_1p^{x_1} = -\ln(p)t_2p^{x_2} = \cdots = -\ln(p)t_kp^{x_k} \\
		\implies& t_1p^{x_1} = t_2p^{x_2} = \cdots = t_kp^{x_k} \\
		\implies& \log_p(t_1) + x_1 = \log_p(t_2) + x_2 = \cdots = Z.
	\end{align*}
	Summing over all of these equivalent expressions gives
	\begin{align*}
		kZ &= \sum_{i = 1}^k \log_p(t_i) + \sum_{i = 1}^k x_i \\
		\implies Z &= \frac1k \Bigg(\sum_{i = 1}^k \log_p(t_i) + N\Bigg).
	\end{align*}
	Having obtained this constant $Z$, we can now immediately calculate the solution via
	
	$$x_i = Z - \log_p(t_i)$$

	In most cases, this method yields a valid solution. However, we have not yet fully enforced our constraints. If $T$ contains a particularly wide range of values, then the assignment produced may actually have a negative number of agents assigned to one or more of the least valuable tasks. If this is the case, we remove the problematic tasks from $T$ and recompute with the same method, repeating this process until all values are non-negative. This is equivalent to optimizing over the boundary formed by setting these values identically to 0. Due to the convexity of $f$, this iterative procedure will always yield the global maximum subject to our constraints.

	Now, to convert this solution back into the discrete case, we simply take the floor of each value, calculate the number of remaining unassigned agents, and then use the greedy procedure from Algorithm 1 to assign them sequentially. Algorithm 2 demonstrates this approach.

	\begin{algorithm}[h]
		\caption{Relaxed solution to Problem 1}
		\begin{algorithmic}
			\STATE $x \gets [0, 0, \dots, 0]$ \COMMENT{Solve continuous relaxation}
			\STATE $C \gets 0$
			\REPEAT
				\STATE $Z \gets \frac1k \big(\sum_{i = 1}^k \log_p(t_i) + N\big)$
				\STATE $c \gets 0$
				\FOR{$i = 1$ to $k$}
					\STATE $x_i \gets Z - \log_p(t_i)$
					\IF{$x_i < 0$}
						\STATE $c \gets c + 1$
					\ENDIF
				\ENDFOR
				\STATE $k \gets k - c$
				\STATE $T \gets [t_1, t_2, \cdots, t_k]$
				\STATE $x \gets [x_1, x_2, \cdots, x_k]$
				\STATE $C \gets C + c$
			\UNTIL{$x_i \ge 0$ for all $1 \le i \le k$}
			\FOR[Convert to discrete]{$i = 1$ to $k$}
				\STATE $x_i \gets \lfloor x_i \rfloor$
			\ENDFOR
			\STATE $L \gets N - \sum_{i = 1}^k x_i$
			\FOR{$i = 1$ to $L$}
				\STATE $z \in \argmax_{1 \le j \le k}\big(t_j \cdot p^{x_j}(1 - p)\big)$
				\STATE $x_z \gets x_z + 1$
			\ENDFOR
			\RETURN $[x_1, x_2, \cdots, x_k, \underbrace{0, 0, \cdots, 0}_C]$
		\end{algorithmic}
	\end{algorithm}

    \begin{theorem}[Performance of Algorithm 2]
    Given an instance of Problem 1, the performance of Algorithm 2 is bounded as follows:
    \begin{enumerate}
        \item \emph{(Optimality)} The assignment returned yields the maximum possible expected value for total profit.
        \item \emph{(Runtime)} The overall time complexity is $O(k^2)$, which is independent of the number of agents.
    \end{enumerate}
    \end{theorem}
   It is evident from our derivation that Algorithm 2 is optimal for the relaxed version. The conclusion that it is also optimal for the original version follows from the correctness of Algorithm 1 and is further supported by empirical results. The runtime is explained by the following. Since computing $Z$ and its corresponding assignment is accomplished in $O(k)$, and at most $k$ such instances are required, the continuous relaxation is solved in $O(k^2)$. Further, since $L \le k$, converting to a discrete assignment costs at most $O(k\log k)$. Thus, the overall complexity of Algorithm 2 is $O(k^2)$.

	\section{Adversarial Resilience}
	\subsection{Problem Formulation}
	Problem 1 effectively models scenarios in which we wish to maximize the expected value of our profit. However, this is not always the best metric for success in failure-prone environments. Consider a scenario in which completing a set of tasks is mission-critical. That is, if the total profit gained from attempting to complete these tasks is below some threshold, then the overall mission will fail. In cases like these, it may be helpful to consider the \emph{worst-case} profit of a given assignment as opposed to the \emph{average} profit as we did in Problem 1.

	In stochastic environments, the worst case occurs when all agents fail, resulting in a profit of 0. For this model, we will assume that the maximum number of agents that can fail is capped at some given constant $\alpha \le N$. This allows us to define the following notion of failure tolerance levels. 
	
	\begin{definition}[$\alpha$-tolerance]
	    Given an assignment $x$, a profit threshold $F \in \mathbb{R}$, and a failure limit $\alpha \in \mathbb{Z}$, we say that $x$ is $\alpha$-tolerant iff, for all cases in which at most $\alpha$ agents fail, the resulting profit is at least $F$.
	\end{definition}
	
	 In other words, an assignment is $\alpha$-tolerant if it guarantees profit $F$ even in the worst case. This definition encapsulates the notion of robustness in a way that effectively models many real world scenarios. Two examples are hard drive failures in RAID arrays (where a certain number of drives are required to retain data integrity) \cite{patterson} and cooperative swarm robotics (where a large number of failures can lead to a break in communication) \cite{mohan}. 
	
	An equivalent way of viewing this notion of robustness is to suppose that agents fail, not by chance, but instead by the actions of an intelligent adversary. In particular, we consider scenarios in which first, a defender assigns agents to tasks, and then an attacker disables at most $\alpha$ of the agents with the intention of minimizing the defenders profit. The attacker has full knowledge of the defender's assignment ($x$) and the defender's valuations of the tasks ($T$). The targets selected by the attacker constitutes the worst-case failure for the given assignment. As such, these two formulations are completely equivalent. Any assignment that yields profit at least $F$ in the adversarial environment is clearly $\alpha$-tolerant in the stochastic environment (and vice-versa). This perspective also lends itself to various real-world scenarios, the most obvious of which being military applications \cite{young}.
	
	We begin by formalizing the set of valid target selections available to the attacker in response to assignment $x$ with failure limit $\alpha$:
	$$\Delta_{\alpha}(x) = \Big\{\delta \in \mathbb{Z}^k_{\ge 0} \Bigm\vert \Big(\textstyle{\sum_{i=1}^k \delta_i \le \alpha} \Big) \land \Big(\forall_{1 \le i \le k} (\delta_i \le x_i) \Big) \Big\}$$
	Selection $\delta \in \Delta_{\alpha}(x)$ represents the attack that disables $\delta_i$ of the agents assigned to task $i$. Thus, $x - \delta$ gives the agents remaining after the attack has occurred. When an attacker behaves optimally, the $\delta$ selected will result in the minimum profit for the defender. The problem of finding the best assignment in the face of an optimal attacker is modeled by the following.

	\begin{problem}[Adversarial Failure]
		Given task values $T \in \mathbb{R}^k$, number of agents $N \in \mathbb{Z}_{\ge 0}$, and failure limit $\alpha \in \mathbb{Z}_{\ge 0}$ (with $\alpha \le N$), find an assignment $x \in \mathbb{Z}^k$ that
		\begin{align*}
			\text{maximizes\quad} &\min_{\delta \in \Delta_{\alpha}(x)} \sum_{i=1}^k t_i \cdot H(x_i - \delta_i) \\
			\text{subject to\quad} &\sum_{i = 1}^{k} x_i \le N \text{\quad and\quad} x_i \ge 0 \hspace{0.8mm}\text{ for all }\hspace{0.8mm} 1 \le i \le k.
		\end{align*} 
	\end{problem}

	Solving the inner minimization problem reflects the role of the optimal attacker, and the overall maximin problem the role of the defender. Problem 2 may be interpreted as a two-stage perfect information zero-sum sequential game (sometimes called a ``Stackelberg" game) between these two players \cite{myerson}. Consider the following example. Given $N = 9$, $T = [90, 65, 55, 30, 15]$, $\alpha = 3$, the defender should implement the assignment $x^* = [3, 2, 2, 1, 1]$. In response, the attacker will implement $\delta = [0, 2, 0, 1, 0]$, resulting in a total profit of 160.
	
	Again, we denote the objective function $f$. The decision version of Problem 2 is to determine if there exists an assignment $x$ such that $f(x) \ge F$ for a given $F \in \mathbb{R}$. This corresponds directly to asking if there exists an $\alpha$-tolerant assignment under profit threshold $F$. 
	
	\subsection{Exact Solution}
	We begin by fully characterizing the behavior of the optimal attacker in response to assignment $x$. Observe that the attacker only serves to gain when they target all agents assigned to a given task. Thus, the attacker's problem can be simplified to that of selecting \emph{tasks} instead of agents. Specifically, the attacker wishes to select a subset of tasks with maximal total value, such that the sum of the agents assigned to those tasks is at most $\alpha$ (thus maximizing the harm to the defender). This is an instance of the 0-1 Knapsack Optimization Problem, where the size of the knapsack is $\alpha$, the weight of item $i$ is $x_i$, and the value of item $i$ is $t_i$. This is itself an NP-Hard problem. However, it omits a dynamic programming solution which can be implemented in pseudo-polynomial time: $O(\alpha k)$.

	With this in mind, it is now possible to implement a brute force solution that iterates over every valid assignment, solving the knapsack instance for each one, and returning the assignment that yields the greatest profit. The runtime of this approach is given by:
	$$O(\alpha k) \cdot |X| = O\bigg(\frac{Nk(N + k)!}{N!k!}\bigg)$$
	where $X$ denotes the set of all valid assignments. The cardinality of $X$ follows from the observation that $X$ represents the number of compositions of $N$ into at most $k$ parts. This can be greatly improved by the following observation. 

	\begin{proposition}[Decreasing Optimal Assignment]
		For any instance of Problem 2, there exists an optimal assignment for which the agents are distributed in decreasing order. Formally, $\exists \tilde{x} \in X$ s.t. $f(\tilde{x}) = \max_{x \in X} f(x)$ and $\tilde{x}_i \ge \tilde{x}_j$ for all $1 \le i < j \le k$.
	\end{proposition}

	We now restrict the search space to only the assignments that that satisfy this decreasing property. Denote this smaller set $\tilde{X}$. Proposition 1 guarantees that $\argmax_{x \in \tilde{X}}$ is an optimal solution to Problem 2. In contrast with $X$, $\tilde{X}$ represents the \emph{partitions} of $N$ into at most $k$ parts, which has a much lower asymptotic bound \cite[Eq.~26.9.10]{NIST:DLMF}. The runtime of brute force is now given by:

	$$O(\alpha k) \cdot |\tilde{X}| = O\bigg(\frac{N^k}{(k-1)!^2}\bigg)$$

	Clearly, this is still only suitable for small values of $N$ and $k$. However, it is unlikely that a significantly faster exact solution can be found. A branch-and-bound approach may offer some improvements but is greatly limited by the highly non-convex nature of the objective function. In the following sections, we show that both Problem 2 and its decision version are strictly NP-Hard, and then provide an approximate solution.

	\subsection{NP-Hardness}
	It is easy to see that the decision version of Problem 2 is strictly NP-Hard. Any algorithm that would be able to verify if a given assignment achieves profit at least $F$ in polynomial time would also be able to solve the Knapsack Decision Problem (KDP) in polynomial time. This follows precisely from the behavior of the attacker as discussed in the previous section. However, it is well known that KDP is NP-Complete \cite{karp}. Thus, unless $P=NP$, a solution to the decision version cannot be verified in polynomial time, making it strictly NP-Hard. Since the optimization version of Problem 2 is at least as hard as the decision version, it is also strictly NP-Hard.

	\subsection{Approximate Solutions}
	Our approximation approach involves sampling $\tilde{X}$ by selecting a subset of plausible assignments, calculating the resulting post-attack profit for each one (by solving the inner minimization problem), and returning the assignment which yields the maximum. We restrict the number of assignments sampled to be polynomial in $k$. In particular, we only sample the assignments which \emph{evenly} distribute all $N$ agents amongst the $m$ most valuable tasks for some $1 \le m \le k$. When $N$ is not evenly divisible by $m$, we give the most valuable tasks precedence in receiving the additional agents. Formally, for every $m$, we set $c = \lfloor N/m \rfloor$ and sample the assignment
	$$x_i = \begin{cases}
		c + 1 &1 \le i \le N - cm\\
		c &N - cm < i \le m \\
		0 & m < i \le k
	\end{cases}$$
	For example, $m = 2$ gives $\big[\frac{N}{2}, \frac{N}{2}, 0, 0, \cdots, 0\big]$ when $N$ is even and $\big[\frac{N + 1}{2}, \frac{N - 1}{2}, 0, 0, \cdots, 0\big]$ when $N$ is odd. Restricting our search space in this manner allows us to find adequate solutions while avoiding the combinatorial intractability of $\tilde{X}$. However, one potential problem is that we must solve the inner minimization problem for each sample. As shown before, this reduces to the knapsack optimization problem in the general case, which can be costly even when using dynamic programming. However, since the assignments in question are of a specific form, we can actually solve these knapsack instances exactly in polynomial time. The key insight is that the number of agents assigned to each task is limited to one of three values: $0$, $c$, or $c + 1$. This greatly reduces the number of feasible attacks and allows us to efficiently iterate over the boundary of the space.
	
	Algorithm 3 utilizes this approach to find an approximate solution in polynomial time. For each of the $k$ different values of $m$, the construction and evaluation of $x$ are both $O(k)$. Thus, the overall runtime of Algorithm 3 is $O(k^2)$. The effectiveness of this algorithm is discussed in the next section.

	\begin{algorithm}[t]
		\caption{Approximate Solution to Problem 2}
		\begin{algorithmic}
			\STATE $U_0 \gets 0$ \COMMENT{Construct prefix sum array}
			\FOR{$i = 0$ to $k$}
				\STATE $U_{i + 1} \gets U_i + t_i$
			\ENDFOR
			\STATE $p_{max} \gets -\infty$
			\FOR{$m = 1$ to $k$}
				\STATE $x \gets [0, 0, \cdots, 0]^{\top}$ \COMMENT{Construct assignment}
				\STATE $c \gets \lfloor N/m \rfloor$
				\STATE $d \gets N - cm$
				\FOR{$i = 1$ to $m$}
					\STATE $x_i \gets c$
					\IF{$i \le d$}
						\STATE $x_i \gets x_i + 1$
					\ENDIF
				\ENDFOR
				\STATE $p_{min} \gets \infty$ \COMMENT{Evaluate assignment}
				\FOR{$r = 0$ to $\min\big(d, \big\lfloor \frac{\alpha}{c + 1} \big\rfloor\big)$}
					\STATE $s \gets \min\big(m - d, \big\lfloor \frac{\alpha - r(c + 1)}{c} \big\rfloor \big)$
					\STATE $p \gets (U_d - U_r) + (U_m - U_{d + s})$
					\STATE $p_{min} \gets \min(p_{min}, p)$
				\ENDFOR
				\IF{$p > p_{max}$}
					\STATE $p_{max} \gets p$
					\STATE $x_{max} \gets x$
				\ENDIF
			\ENDFOR
			\RETURN $x_{max}$
		\end{algorithmic}
	\end{algorithm}

	\subsection{Empirical Results}
	In order to determine the effectiveness of our approximation scheme, Algorithm 3 was run against a large number of test cases (i.e., instances of Problem 2). For each case, the optimal solution was computed via the brute-force method described in section 3.2 and compared to the approximate by taking the ratio of their respective post-attack profits. Two additional baseline solutions were also computed: the ``greedy" baseline, which naively assigns $\alpha + 1$ agents to as many of the tasks as possible (in the order of decreasing profits), randomly assigning any remaining agents; and the ``expectation" baseline, which treats the problem as if it were stochastic, calculating $p$ as $\alpha / N$ and then applying Algorithm 2.

	Three distinct test suites were run, each with a different model for the distribution of task values. Specifically, the distributions tested were the continuous distribution on $[0, 1]$, the exponential distribution with $\lambda = 2$, and the beta distribution with $\alpha = 6$, $\beta = 2$. Respectively, these represent scenarios in which the task values are skewed symmetrically, positively, and negatively \cite{forsyth}. For each distribution, 10,000 trials were run, and the values for the number of tasks, number of agents, and failure limit were randomly selected from $\{(k, N, \alpha) \in \mathbb{Z}^3 \mid 2 \le k \le N \le 30, \hspace{1mm} 2 < \alpha < N\}$. 
	
    Figure 1 shows the results of these tests, displaying both the average and minimum observed profit ratios for each type. Table 1 shows the average runtime of each algorithm over all three test suites. Evidently, Algorithm 3 achieved an average of over 95\% optimal in all three cases, with performance as high as 99\% in the case of the beta distribution. This significantly outperforms both baselines. Further, all 30,000 test cases achieved at least 70\% optimal, implying that Algorithm 3 has reliable worst-case performance. We hypothesize that Algorithm 3 will always achieve at least $\frac23$ optimal but have not yet been able to formally prove this conjecture.
	
	\begin{figure}[t]
	    \centering
	    \includegraphics[width=0.75\linewidth]{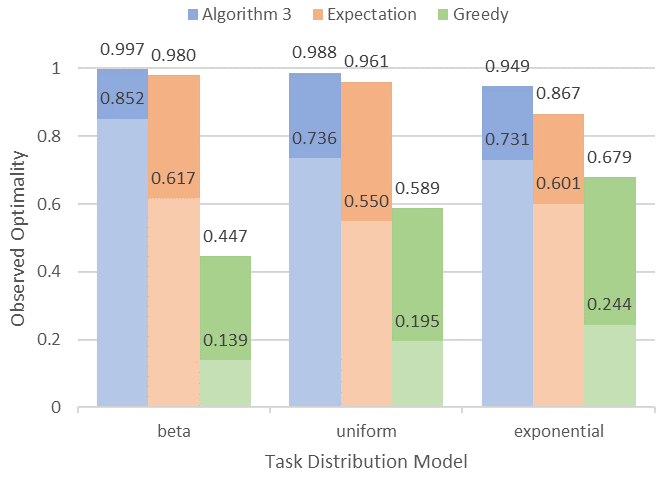}
	    \caption{Optimality of Algorithm 3 compared to baselines under various distributions. Both average and minimum observed ratios are shown for each case.}
	    \label{fig:empirical_results}
    \end{figure}

    \begin{table}[]
		\centering
		\vspace{-2mm}
        \begin{tabular}{|l|l|}
            \hline
            \textbf{Algorithm} & \textbf{Runtime (ms)} \\ \hline
            Brute Force        & 453.1                         \\ \hline
            Algorithm 3        & 0.176                         \\ \hline
            Expectation        & 0.176                         \\ \hline
            Greedy             & 0.017                         \\ \hline
        \end{tabular}
        \caption*{Table 1: Average runtimes of exact, approximate, and baseline algorithms. Tests were limited to cases with 30 or fewer agents. Run in Python 3 on an Intel i7-4790k @ 4.5GHz.}
        \vspace{-5mm}
    \end{table}

	\section{Conclusion and Future Work}
    We take key steps towards ensuring robustness in multi-agent systems that suffer from failures. In particular, we formulate the Robust Multi-Agent Assignment Problem for stochastic and adversarial environments within a set framework and demonstrate the need for such models. We provide an efficient exact solution for the stochastic case and an approximate solution for the adversarial. We demonstrate and analyze the effectiveness of these algorithms with custom simulations and discuss guaranteed performance. Notably, the results of this paper extend to any multi-agent system that fits the framework, not just those that arise in robotics. 

    This work is still ongoing, with our primary focus placed on increasing the complexity of the model to account for a broader class of problems. For example, this might include having a different probability of failure associated with each task. The framework could also be expanded in various ways to include heterogeneous agents. For example, suppose that each agent has a certain proficiency which uniquely determines it's probability of failure. The notion of robustness in multi-agent systems is both potentially important and largely unexplored, leading to many possible avenues for future research.
	
	\section*{Acknowledgments}
	This work is supported, in part, by the National Science Foundation under Grant No. 1943368.

    \bibliographystyle{IEEEtran}
    \bibliography{IEEEabrv, refs}

\end{document}